\pdfoutput=1

\documentclass[11pt]{article}

\usepackage[]{acl}

\usepackage{times}
\usepackage{latexsym}

\usepackage[T1]{fontenc}

\usepackage[utf8]{inputenc}

\usepackage{microtype}

\usepackage{caption}
\usepackage{subcaption}
\usepackage{booktabs}
\usepackage{graphicx}
\usepackage{amsmath}
\usepackage{empheq}

\title{Effectiveness of Mining Audio and Text Pairs from Public Data \\ for Improving ASR Systems for Low-Resource Languages}
\author{
        Kaushal Santosh Bhogale$^{\lambda\psi*}$  \hspace{0.2cm} Abhigyan Raman$^{\psi}$\thanks{* The first two authors have contributed equally.} \hspace{0.2cm} Tahir Javed$^{\lambda\psi}$  \hspace{0.2cm} \\
        \textbf{Sumanth Doddapaneni}$^{\lambda\psi}$ \hspace{0.2cm} \textbf{Anoop Kunchukuttan}$^{\psi\mathsection}$ \\ 
        \textbf{Pratyush Kumar}$^{\psi\mathsection}$ \hspace{0.2cm}
         \textbf{Mitesh M. Khapra}$^{\lambda\psi}\thanks{$\dag$ Corresponding author: miteshk@cse.iitm.ac.in}$ 
    \\ \\
    $^\lambda$Indian Institute of Technology, Madras \\
    $^\psi$AI4Bharat \hspace{0.2cm} $^\mathsection$Microsoft \\
}

\begin{document}
\maketitle
\begin{abstract}
End-to-end (E2E) models have become the default choice for state-of-the-art speech recognition systems. 
Such models are trained on large amounts of labelled data, which are often not available for low-resource languages. 
Techniques such as self-supervised learning and transfer learning hold promise, but have not yet been effective in training accurate models.
On the other hand, collecting labelled datasets on a diverse set of domains and speakers is very expensive. 
In this work, we demonstrate an inexpensive and effective alternative to these approaches by ``mining'' text and audio pairs for Indian languages from public sources, specifically from the public archives of All India Radio. 
As a key component, we adapt the Needleman-Wunsch algorithm to align sentences with corresponding audio segments given a long audio and a PDF of its transcript, while being robust to errors due to OCR, extraneous text, and non-transcribed speech. 
We thus create Shrutilipi, a dataset which contains over 6,400 hours of labelled audio across 12 Indian languages totalling to 4.95M sentences. 
On average, Shrutilipi results in a 2.3$\times$ increase over publicly available labelled data.
We establish the quality of Shrutilipi with 21 human evaluators across the 12 languages. 
We also establish the diversity of Shrutilipi in terms of represented regions, speakers, and mentioned named entities. 
Significantly, we show that adding Shrutilipi to the training set of Wav2Vec models leads to an average decrease in WER of 5.8\% for 7 languages on the IndicSUPERB benchmark. 
For Hindi, which has the most benchmarks (7), the average WER falls from 18.8\% to 13.5\%. 
This improvement extends to efficient models: We show a 2.3\% drop in WER for a Conformer model (10$\times$ smaller than Wav2Vec).
Finally, we demonstrate the diversity of Shrutilipi by showing that the model trained with it is more robust to noisy input. 
\end{abstract}

\section{Introduction}

Current state-of-the-art speech recognition systems often employ end-to-end (E2E) models \cite{li2022recent,graves2012sequence,graves2012connectionist,DBLP:conf/interspeech/SoltauLS17,gulati2020conformer,babu2021xls} which combine acoustic, pronunciation, and language models into a single network.
Such models are often large (order millions of parameters) and require compute-heavy training on large datasets of labelled audio. 
While reducing the word-error-rate (WER) of high-resource languages such as English, such models increase the performance gap of speech systems for low-resource languages, further disadvantaging adoption of AI models for low-resource languages.

A robust approach to address this gap is to collect labelled datasets for low-resource languages.
However, creation of high quality datasets can be expensive given the logistics of collecting data across a large diversity of languages and dialects \cite{gumperz1961speech}.
Another approach to reduce the gap between languages is self-supervised learning.
Models such as Wav2Vec \cite{baevski2020wav2vec} can be pretrained on large easier-to-obtain unlabelled datasets and then fine-tuned with smaller labelled datasets.
This was demonstrated for Indian languages in \citet{javed2022towards} with pretraining on 40 languages and fine-tuned models for 9 languages. 
However, the WER reported for Indian languages is still much higher than what is achieved with equivalent models for high-resource languages. 
Another approach is cross-lingual transfer of knowledge from high to low-resource languages. \citet{Scharenborg2017BuildingAA}
Specifically, labelled datasets for high-resource languages can be \textit{transliterated} to low-resource languages, similar to how language understanding tasks are created for low-resource languages by translation \cite{khare2021low}. 
However, the resultant accuracy still leaves large WER gaps w.r.t. high-resource languages \cite{khare2021low}. 
Further, transfer learning is dependent on non-native speakers and content which may cause robustness issues on more diverse benchmarks.
Thus, while self-supervised and transfer learning can effectively reduce WER, they still leave a role for larger datasets to further reduce the large gap to high-resource languages.

\begin{figure}[t!]
    \centering
    \includegraphics[width=0.8\columnwidth]{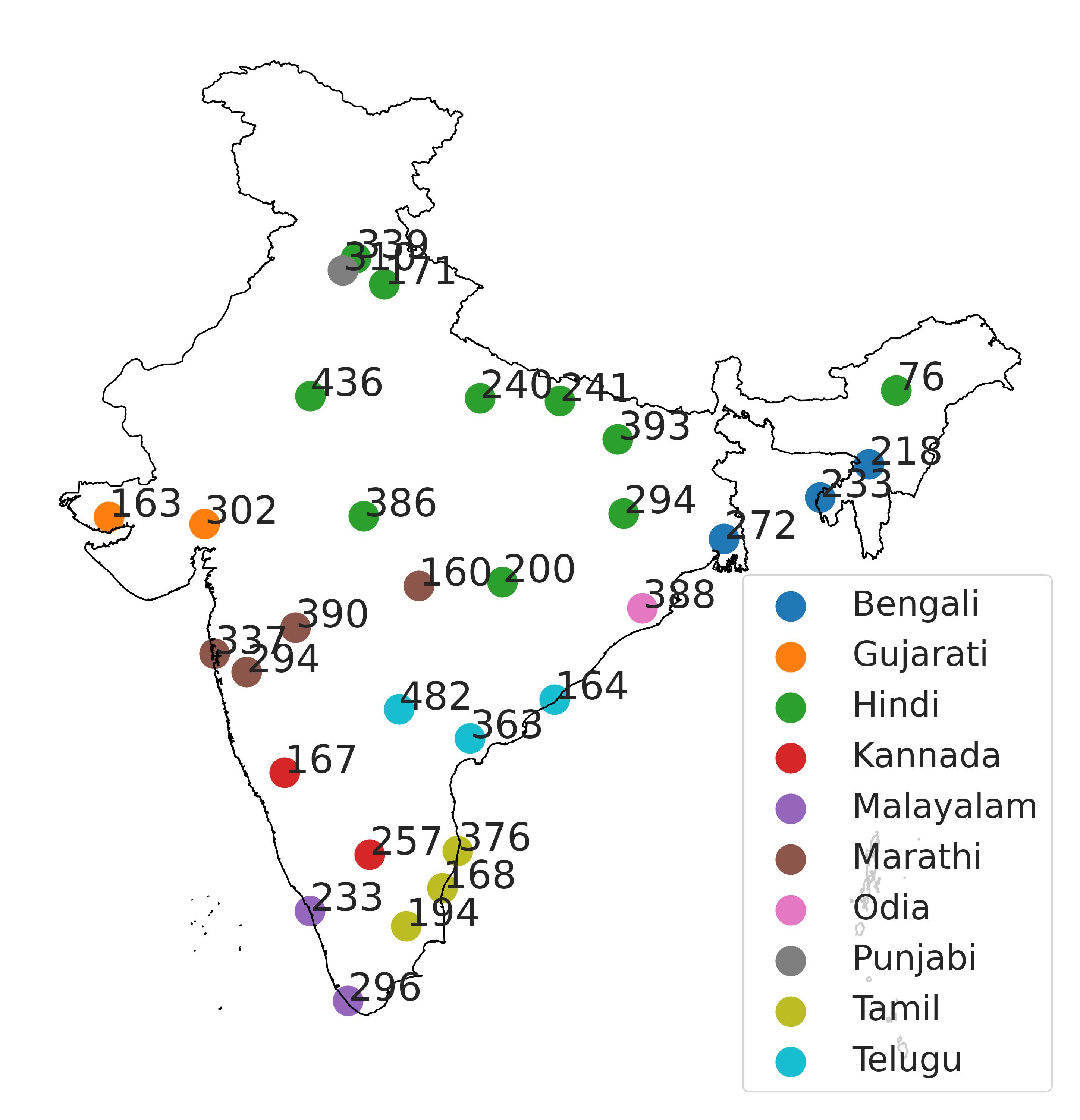}
    \caption{No. of hours of data obtained from Regional radio stations of All India Radio}
    \label{fig:india-map}
\end{figure}

In this paper, we propose an alternative technique to create diverse datasets for low-resource languages. 
Our approach is motivated by a recent work in Neural Machine Translation (NMT) for Indian languages \cite{ramesh2022samanantar}, where billions of sentences in Indian languages  were collected from public sources, and then pairs of sentences with high semantic similarity were mined. 
The obtained \textit{bi-text pairs} were used to train NMT systems which improved the accuracy by several BLEU \cite{papineni-etal-2002-bleu} points for 12 Indian languages, in spite of inexact alignment of the pairs.
Similarly, in this paper, we mine audio segments and corresponding text segments for Indian languages from publicly available sources. 
Specifically, we process the archives of India's public broadcaster, All India Radio (AIR), which broadcasts radio bulletins in various Indian languages from 45 stations from February 2018. 
The geographical distribution of these stations and the number of hours of archival audio data are shown in Figure~\ref{fig:india-map}. 
In 12 Indian languages, the archive has over 9,700 hours of audio and is recognized as a major source of Indian language data \cite{giz}.

While both audio bulletins and PDF transcripts are available from AIR, extracting audio and corresponding text pairs, say at a sentence granularity, is not straightforward.
We extracted text from the PDF documents with OCR which was error-prone given use of proprietary non-UTF font encodings, presence of formatted content with formats varying from one station to other, presence of metadata such as speaker cues, etc.
Audio bulletins too had challenges such as presence of intro and outro segments often containing music, non-transcribed speech such as announcements and external news clips, presence of background music, and code-mixed speech.
Given these irregularities, a core requirement is a robust and systematic approach to align document-scale audio and text segments across languages and programs.

We propose an alignment methodology with three steps. 
First, we process the audio bulletin with an ASR system (IndicWav2Vec \cite{javed2022towards}) to obtain frame level emissions which are collapsed into a character sequence using CTC alignment.
Similarly, we pass the PDF documents through OCR and text cleanup to obtain the transcription.
Second, we adapt the Needleman-Wunsch algorithm \cite{needleman1970general}, which was originally designed to align protein or nucleotide sequences, to align the outputs of ASR and text extraction.
Through experimentation, we set the parameters of the algorithm to allow for insertions and deletion of characters. 
Importantly, the alignment is at document-scale, which on average is about 8 mins long with tens of thousands of characters.
Third, we segment the extracted text by sentence boundaries and the audio bulletin by the time intervals based on alignments identified using CTC \cite{graves2012connectionist} and Needleman-Wunsch algorithm \cite{needleman1970general}.
We retain each pair of segmented text and audio if a defined Levenstein distance similarity ratio is above a chosen threshold $\tau$. 
Thus, the output of alignment is a sequence of audio and text pairs corresponding to sentences of text. 

We apply these methods to the AIR archive to obtain the Shrutilipi (In Sanskrit, Shruti means 'Sound' and Lipi means 'script'.) dataset with 6,457 hours of data at a value of $\tau = 0.8$, which is more than 60\% of the raw audio. 
The corresponding number at $\tau = 0.95$ is 3,239 hours.
We evaluate the quality of Shrutilipi with 4,050 pairs graded by 21 annotators.
The annotators found quality of alignment to be consistently high for $\tau = 0.95$ across stations. 
Further, the annotators localized a majority of alignment errors either at the start or end of segments. 
We show that training of ASR systems is more forgiving of such errors.
We report on the speaker and content diversity of Shrutilipi.
For speaker diversity, we obtain speaker-related embeddings by training a speaker verification task.
These embeddings are significantly more diverse for Shrutilipi w.r.t. the MUCS dataset.
For content diversity, we find that Shrutilipi has significantly more unique references to named entities across categories w.r.t. to the MUCS dataset. 

We evaluate the value of Shrutilipi by training ASR systems for 7 Indian languages using the Wav2Vec  architecture \cite{baevski2020wav2vec} with existing baselines \cite{javed2022towards}. 
On the IndicSUPERB benchmark \cite{javed2022indicsuperb}, we show that addition of Shrutilipi to the training dataset of Wav2Vec decreases WER averaged across 7 languages by 5.82\%. 
For Hindi, where we have 7 public benchmarks, the average WER falls from 18.8\% to 13.2\%.
This observed reduction is on top of improvements made in \citet{javed2022towards} of pretraining, thus demonstrating that gains with mining data can compose with those from self-supervised learning.
The improvement in WER extends to efficient models: We show a 2.26\% reduction for a Conformer \cite{gulati2020conformer} model which is 10$\times$ smaller than Wav2Vec. 
The above results are on the entire Shrutilipi data created for $\tau = 0.8$.
For Hindi, we compare accuracy of models for different values of $\tau = 0.8, 0.9, 0.95$ and found that WER was least for $\tau = 0.8$.
Finally, we create a hard ASR benchmark for Hindi by introducing background noise, and show that training with Shrutilipi leads to a lower increase in WER. 

In summary, we propose a methodology to perform long-audio alignment and apply it to the public archives of AIR to create Shrutilipi. 
We demonstrate the quality and utility of Shrutilipi with human evaluations, and WER reductions of ASR models on public and noisy benchmarks.
We hope that this template can be replicated with other sources of data for Indian languages, and for other languages around the world, to bridge the widening gap between ASR systems for high and low-resource languages.


\begin{figure*}[ht!]
    \centering
    \includegraphics[width=\textwidth]{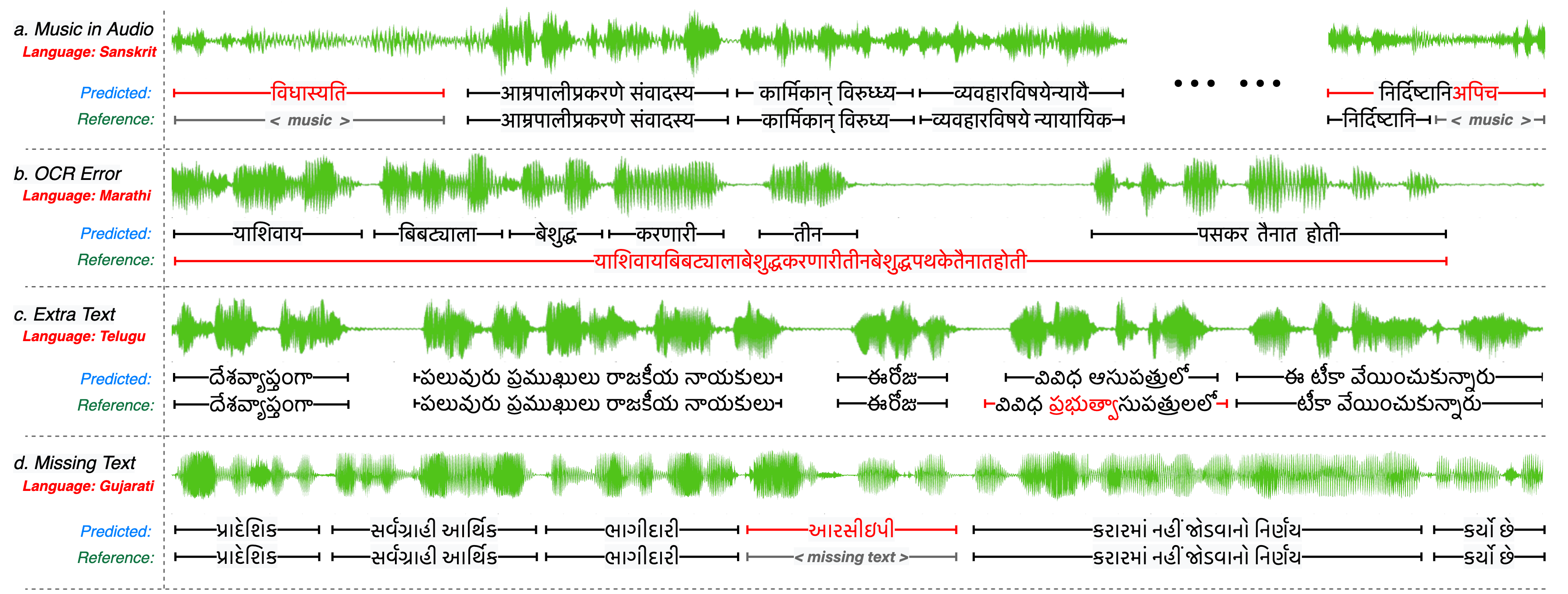}
    \caption{Types of irregularities in audio inputs and transcripts which make mining audio and text segments challenging.}
    \label{fig:align-error}
\end{figure*}

\section{All India Radio Dataset} \label{section2}

Though the methodology presented in this work applies generally in mining audio and text pairs for long audio, we will specifically focus on the dataset available from All India Radio (AIR).
In this section, we detail this dataset, its language-wise statistics, and some of the  issues that make mining text and audio pairs non-trivial. 

AIR is the national radio broadcaster of India, a Prasar Bharati
division, that streams radio programs in all major Indian languages. 
The AIR website\footnote{http://newsonair.in/} hosts thousands of hours of audio and PDF transcripts for programs.
In this paper, we work with data for 12 Indian languages, collectively representing 1.1B number of speakers in the Indian subcontinent.
There are multiple stations creating content in each language categorised into two types - the News Services Division and Regional News. 
The stations are geographically scattered across India (see Figure \ref{fig:india-map}), indicating diversity in regional representation.
The audio data comes from news bulletins that are aired either daily or bi-daily depending on the station. 
The bulletins vary in length, but are typically 5 min, 10 min, or 15 min in duration. 
The audio data is in MP3 format sampled at 44KHz and is either mono-channel or stereo. 
We standardize the data by converting it into 16KHz WAV mono-channel format.
The transcriptions for each bulletin are available as PDF documents. 
Almost every radio station has a different document style.

We collect bulletins aired between 21 Feb 2018 and 17 May 2021.
After extraction, we filter out bulletins which have corrupted audio or transcript files.
After this basic filter, we collect a total of 72,580 bulletins with 9,695 hours of audio, averaging about 8 mins per bulletin. We show the detailed statistics of the All India Radio dataset in Table 1.

\begin{table}[]
\centering
\small
\begin{tabular}{ccccc}
\toprule
\textbf{Lang.} & \textbf{Stations} & \textbf{Bulletins} & \textbf{Hours} \\
\midrule
bn & 4  & 5.7K & 0.64K \\
gu & 3  & 5.6K & 0.68K \\
kn & 3  & 4.9K & 0.65K \\
hi & 11  & 18.8K & 2.35K \\
ml & 3  & 5.7K & 0.87K \\
mr & 5  & 9.0K & 1.28K \\
or & 2  & 5.8K & 0.77K \\
pa & 1  & 0.9K & 0.12K \\
sa & 1  & 0.7K & 0.06K \\
ta & 4  & 7.0K & 1.12K \\
te & 4  & 5.5K & 0.84K \\
ur & 4  & 2.9K & 0.31K \\
\midrule
all & 45 & 72.6K & 9.70K \\
\bottomrule
\end{tabular}
\caption{Statistics of data available from All India Radio archives.}
\label{tab:data}
\end{table}

\subsection{Challenges in Mining Data}

The aim of our work is to process document-scale data to mine audio and text pairs at the sentence-level.
We encountered several irregularities in the datasets which make this mining challenging.
We detail these irregularities to motivate the mining methodology discussed in the next section.

\subsubsection{Audio data}
We found the following irregularities: 
\begin{enumerate}
    \item Bulletins usually contain long \textbf{intro and outro} segments containing music.
    The length and type of music played varies from station to station. 
    We visualize this with an example in Figure~\ref{fig:align-error}(a) for a Sanskrit program.
    \item Bulletins contain \textbf{short non-transcribed speech}, such as speakers introducing themselves, reading titles of the broadcast, or social media handles of AIR.
    \item Bulletins also contain \textbf{long non-transcribed speech}, such as announcements and external news clips such as response of a public figure in a press conference.
    \item Some bulletins contain a \textbf{background music} throughout. 
    \item Many bulletins contain \textbf{code-mixed data} with mainly English words spoken along with the regional language.
\end{enumerate}

The above irregularities impose a few constraints. 
The presence of background music makes it harder to split audio based on voice activity.
Code-mixed data requires support from ASR system being used for mining.
The first three irregularities require the alignment procedure to be able to skip audio segments with no corresponding transcripts. 

\begin{figure*}[h]
    \centering
    \includegraphics[width=\textwidth]{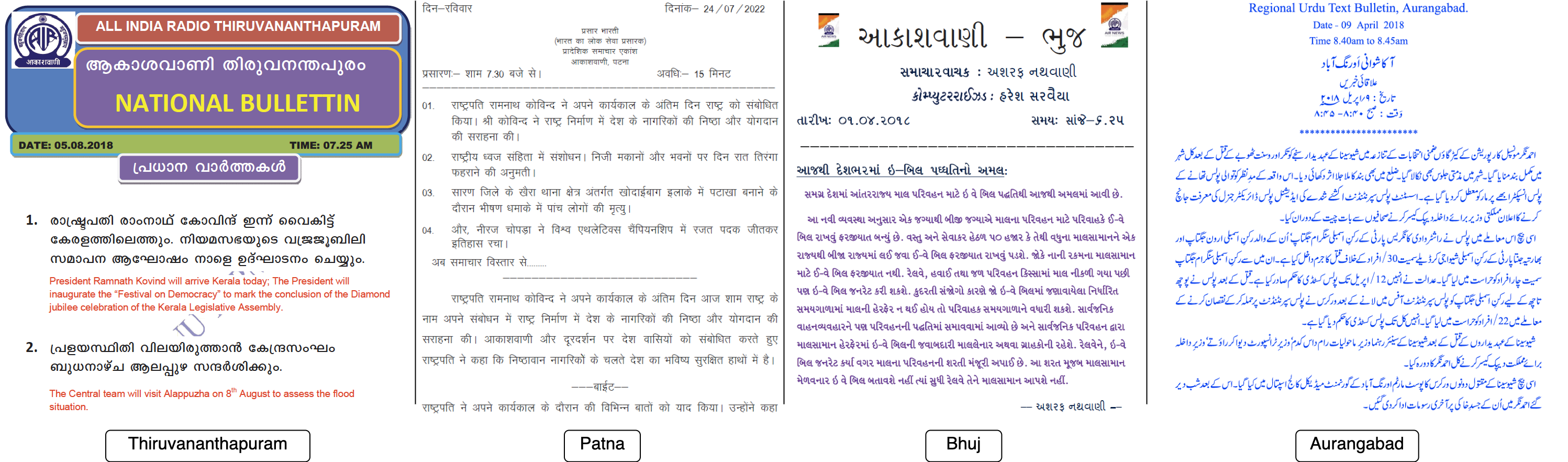}
    \caption{The document style of PDF documents for 4 different stations across languages (Malayalam, Hindi, Gujarati, Urdu). These samples demonstrate the challenges in extracting text like translated text, headers, footers, larger gaps between words due to justified alignment, watermarks, and extraneous text.}
    \label{fig:ocr-diversity}
\end{figure*}

\subsubsection{Transcripts}
We found the following irregularities:
\begin{enumerate}
    \item Most of the transcripts contain \textbf{proprietary encodings} (non-UTF8) due to legacy issues.
    As a result, standard PDF parsers are not effective in extracting text.
    \item The \textbf{custom formats} of the PDFs vary widely across stations and languages.
    These include formatting artefacts such as watermarks, header and footer content.
    \item Most transcripts contain \textbf{extraneous text} such as bulletin headers and section headers which are often not spoken.
    \item Some of the PDF documents contain \textbf{English translations} of the content which are also not spoken. 
\end{enumerate}

Again these irregularities impose constraints.
The first two points require accurate OCR that is robust to format variations and watermarks.
OCR for Indian languages trails others languages, and indeed we observe many text extraction errors.
In Figure~\ref{fig:align-error}(b) we show an example for Marathi where characters were joined into a single word due to reduced spacing between words.
The next two points require alignment method to skip text regions which are not spoken. 

In Figure \ref{fig:ocr-diversity}, we show that the document style of the PDF documents varies significantly across stations. 
We observe the following challenges in each of the documents - (i) Thiruvananthapuram document has a bulletin header, translated text and watermarks (ii) the Patna document has a bulletin header, section headers and large gaps between words due to text justification (iii) the Bhuj document has a bulletin, section headers and a footer (iv) the Aurangabad document has a bulletin header.

\subsubsection{Other issues} 
In addition to the above systematic irregularities, there are various other non-systematic issues.
A news reader might have skipped speaking a word. 
For instance, the audio shown in Figure~\ref{fig:align-error}(c) in Telugu has ``public hospitals'' in the transcript, but the audio contains only ``hospitals''.
A news reader may also speak additional words.
For instance, the audio shown in Figure~\ref{fig:align-error}(d) in Gujarati has ``RCEP kararma'' in Gujarati while the text only contains ``kararma''. 
The word kararma means agreement and RCEP is a specific agreement amongst ASEAN countries, suggesting that the transcript was updated later to include RCEP but is not reflected in the available document. 
The mining methodology must be robust to these variations.

In summary, the data available from AIR contains valuable and diverse content.
But irregularities in audio and transcripts make mining audio and text pairs non-trivial.

\section{Mining Audio and Text Pairs at Document Scale}

\begin{figure*}[t]
    \centering
    \includegraphics[width=\textwidth]{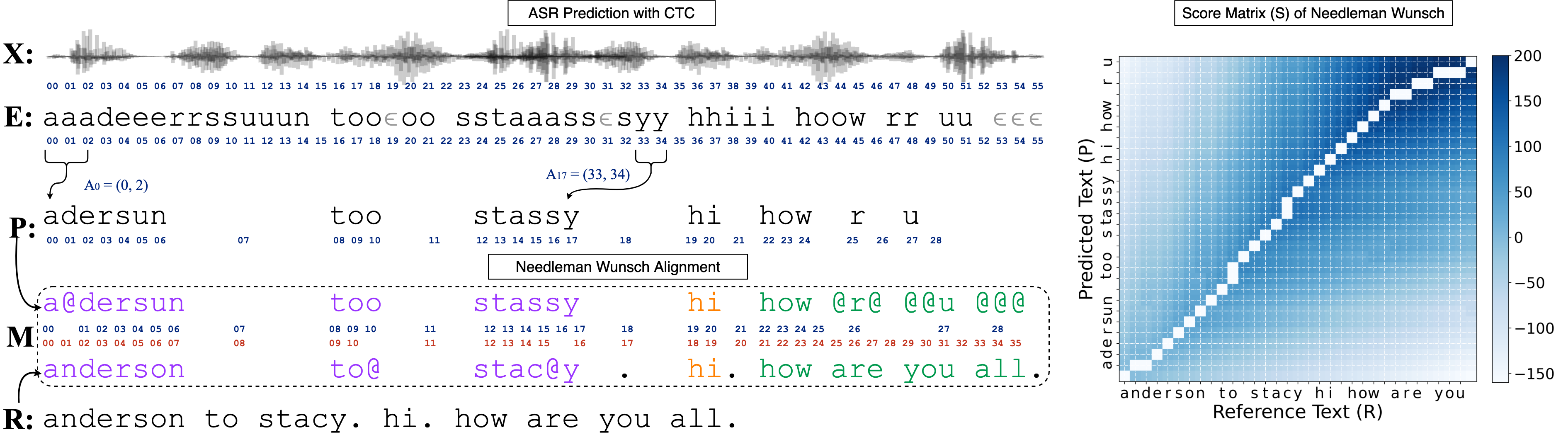}
    \caption{
    Illustration of the alignment algorithm. 
    Audio signal $X$ is processed with ASR to obtain emission sequence $E$, in which repeated emissions are collapsed using CTC to get $P$.
    Start and end indices of CTC alignment are stored in $A$, as show for the first character.
    The text $P$ is aligned with reference text $R$ (with sentence boundaries) using Needleman-Wunsch.
    The algorithm computes a mapping $M$ by finding a score-maximizing path through the score-matrix shown in the right. 
    The mapping can include gaps in either $P$ or $R$ shown by @ which correspond to horizontal or vertical segments in the path, respectively.
    Given $M$, the sentence boundaries are used to find the time-interval in $X$ corresponding to each sentence in $R$.
    }\label{fig:alignment}
\end{figure*}

In this section, we propose a novel technique to mine audio and text pairs at the document scale. 
We describe the alignment technique which consists of 3 main components (i) ASR Prediction using CTC, (ii) Needleman-Wunsch alignment, and (iii) Filtering noisy audio-text pairs. 

\paragraph{Notation} 
The input audio signal $X$ is represented as a sequence $\{x_1, x_2, ..., x_T\}$ of length $T$ where each $x_i$ is an audio frame corresponding to 25ms of audio. 
We assume a ``reference text'' $R$ obtained by text extraction (say through OCR) denoted as a sequence of $N$ characters $\{r_1, r_2, ..., r_{N}\}$ where $r_i$ is a character from a label set $L$ of all valid characters in the language. 
We assume that the reference text can be segmented into sentences and thus define sentence boundaries stored as a sequence $B = \{(\alpha_1, \beta_1), (\alpha_2, \beta_2), ..., (\alpha_W, \beta_W)\}$, where $\alpha_i$ and $\beta_i$ denote the start and end character indices of the sentence in $R$, and $W$ is the number of sentences.
Thus, we have counts along three indices: $T$ for number of audio frames, $N$ for number of characters, and $W$ for number of sentences.
Note that we are working with document scale data, so for an audio signal of 15 mins would have $T = 36,000$ and $N$ as few thousands, and $W$ as few hundreds.
The goal of the mining approach is the following: Find the subset of sentences in $R$, for which an interval of audio frames can be identified whose transcript matches with the sentence.

\subsection{ASR Prediction using CTC}
As the first step, we process the audio signal $X$ through an ASR model which maps each input frame $x_i$ to $L' = L \cup \{blank\}$, to generate emissions $E = (e_1, e_2, ... e_T)$.
Next, we use Connectionist Temporal Classification (CTC) alignment \cite{graves2012connectionist} to collapse repeated characters and remove $blank$ tokens from the emissions to get the predicted sequence of characters $P = (p_1, p_2, ..., p_{N'})$, consisting of $N'$ characters.
We store the CTC alignments, i.e., the start and end indices of emissions that correspond to $p_i$ in a sequence $A = [(\gamma_1, \delta_1), (\gamma_2, \delta_2), ..., (\gamma_J, \delta_{N'})]$ where $\gamma_i$ and $\delta_i$ denote the start and end index respectively. 

The goal of the mining process can now be restated as finding the alignment between the reference text $R$ with $N$ characters and the predicted text $P$ with $N'$ characters.
With such an alignment, we can use sentence boundaries in $B$ and CTC alignments in $A$ to map sentences to audio frames.

\subsection{Needleman-Wunsch Alignment}

We use the Needleman-Wunsch \cite{needleman1970general} algorithm to align predicted text $P$ from an ASR system and a given reference text $R$.
The algorithm uses dynamic programming to align sequences of possibly different lengths accommodating insertions and deletions, as motivated by the problem of finding alignments in proteins.
Given two sequences $R$ and $P$ of sizes $N$ and $N'$ respectively, the goal is to compute a mapping $M$ which is a non-decreasing map of every index $i$ of $R$ to an index $M(i)$ of $P$.
This mapping is computed based on a score matrix $S$ of size $(N + 1) \times (N' + 1)$, where $S_{j,k}$ denotes the alignment score of characters $P[:j]$ and $R[:k]$. 
The algorithm scores pairs of characters with three values: a \textit{Match} score when the two characters exactly match, a \textit{Mismatch} score where the two characters do not match, and a \textit{Gap} score when the chosen alignment involves one character aligning to a gap in the other sequence.
The values chosen for the three scores is application dependent, for instance, we choose value of $+10$, $-5$, and $-5$ for Match, Mismatch, and Gap, respectively, based on empirical evaluation.
Given these character-level scores, dynamic programming is used to find the mapping $M$ that satisfies $M = \arg\min_{O} \sum_{i = 1}^N S_{i, O[i]}$. 

Once we compute $M$, we can obtain alignments of sentences in $R$ to time intervals.
The $i$th sentence in $R$ maps to the character range $[\alpha_i, \beta_i]$ which in turn map to the character range in $P$: $[M(\alpha_i), M(\beta_i)]$, which in turn map to the character range in $E$: $[A(M(\alpha_i)), A(M(\beta_i))]$, which finally map directly to indices in the input audio $X$.
Thus, given a sentence $r$ from $R$, we can compute the corresponding sub-sequence $p$ from $P$, and the sub-sequence $x$ from the input audio $X$. 
We illustrate this with an example in Figure \ref{fig:alignment}.

\subsection{Alignment Score for audio-text pairs}
Since, the Needleman-Wunsch algorithm is a global alignment algorithm, it is possible that it misaligns certain segments of the audio to optimize the scores for other segments. 
Hence, we need a filtering mechanism for extracting high-quality audio-text pairs. 
To address this, we propose an alignment score given by the Levenstein distance similarity ratio between the mined pair of reference sentence $r$ and the predicted sentence $p$, defined as - 
\begin{equation}
    \Delta = 1 - \frac{LD(r, p)}{|r|+|p|}, \quad \Delta \in (0,1)
\end{equation}
where $LD$ is the Levenstein distance between the two strings and $|\cdot|$ denotes the length a sequence. 
We filter out pairs for which $\Delta$ is below a chosen threshold $\tau$. 

In summary, we propose aligning audio at document scale by using Needleman-Wunsch algorithm to align predicted and reference texts, from which we obtain sentence-level segments of audio and text, which are then filtered based on the proposed similarity ratio.

\section{Shrutilipi Dataset} \label{Shrutilipi}

In this section, we discuss applying the mining procedure to the AIR dataset to create  Shrutilipi in 12 languages.
We detail the text processing pipeline, the parameter choices for the alignment algorithm, and statistics of the mined data. 

\subsection{Text processing pipeline}

As discussed, the transcripts in the AIR archives have several irregularities. 
Given the non-standard fonts, we use OCR, specifically Google's Document AI OCR\footnote{https://cloud.google.com/document-ai} which supports all 12 languages that we consider. 
One common OCR error was that text belonging to the same column was treated to be in multiple columns due to large gaps between words.
This changes the order of text extracted and leads to incorrect transcripts.
We correct this by extracting character-wise bounding-boxes and then sequencing characters with a heuristic based on the coordinates, which we describe next.

\paragraph{OCR post-correction}

We propose the following post-correction strategy for OCR. We define a \textit{token} to be a single character detected by the OCR system along with its bounding box information. 
We consider two tokens, say $A$ and $B$, to belong to the same line if the difference in the y-coordinates of the centres of the bounding box of the tokens is less than the height of the first token; and the the difference in the height of the two tokens is less than twice the height of the first token. Formally, the two conditions can be stated as follows -

\begin{align}
\label{eq:correction}
    |A_{centre_{y}} - B_{centre_{y}}| < h(A) \\
    |h(A) - h(B)| < 2 \cdot h(A) \nonumber
\end{align}

where $A_{centre_{y}}$ denotes the y-coordinate of the centre of bounding box of token $A$, and $h(.)$ denotes the height of the bounding box of the token.

We also remove bulletin and section headers which are not spoken, by observing that these headers are often short (less than 5 words) and at the beginning of the document. 
Finally, after the text is extracted, we use a end-of-sentence (EOS) token to segment the text into individual sentences.
In spite of these heuristics, the extracted text contains errors (as we later confirm with human evaluation), and therefore we see value in improving document-level OCR models for Indian languages.

\subsection{Parameters of mining algorithm}
We use Wav2Vec \cite{baevski2020wav2vec} models trained on Kathbath \cite{javed2022indicsuperb} to perform ASR Prediction using CTC. 
We set the match, mismatch, and gap scores of the Needleman-Wunsch alignment to $+10$, $-5$ and $-5$ respectively, based on experiments to maximize the yield of aligned data. 
Based on manual evaluation, we chose the value of $\tau = 0.8$.
While at this threshold aligned pairs have errors, we retain them in anticipation that training an ASR system would benefit from a larger volume of labelled pairs, some of which could be noisy.
We confirm that this is indeed the case in our experiments.

\begin{table}[]
\centering
\small
\begin{tabular}{ccccc}
\toprule
\textbf{Lang.} & \textbf{\# Hrs.} & \textbf{\# Sents.} & \textbf{\# M.W.} & \textbf{\# Y (\%)} \\
\midrule
bn & 0.44K & 0.34M  & 15 & 69 \\
gu & 0.46K & 0.33M  & 15 & 67 \\
hi & 1.62K & 1.10M  & 21 & 69 \\
kn & 0.46K & 0.35M & 12 & 71 \\
ml & 0.36K & 0.59M & 8 & 42 \\
mr & 1.02K & 0.63M  & 15 & 79 \\
or & 0.60K & 0.38M  & 16 & 78 \\
pa & 0.09K & 0.05M  & 26 & 78 \\
sa & 0.02K & 0.03M  & 11 & 42 \\
ta & 0.79K & 0.55M  & 12 & 71 \\
te & 0.39K & 0.43M  & 12 & 47 \\
ur & 0.19K & 0.17M  & 21 & 63 \\
\midrule
Total & 6.46K & 4.95M & 8.90 & 67\\
\bottomrule
\end{tabular}

\caption{Statistics of Shrutilipi dataset (\# M.W.=Mean length of Sentences (in words), \# Y=Yield) }
\label{tab:aligned_data}
\end{table}


\begin{figure}[t]
    \centering
    \includegraphics[width=\columnwidth]{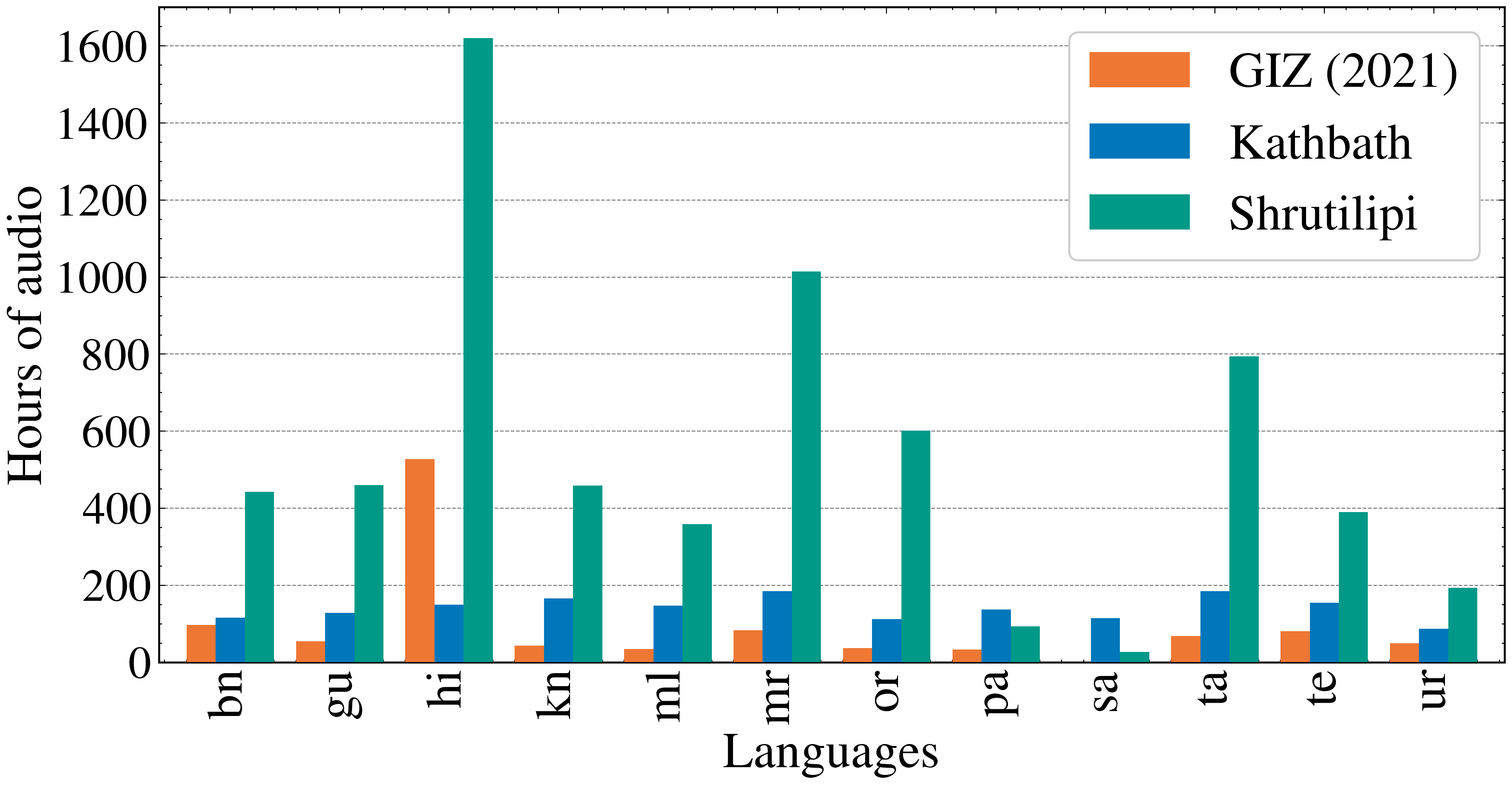}
    \caption{Comparison of Shrutilipi mined data against existing sources}
    \label{fig:existing-data}
\end{figure}

\subsection{Statistics of the Shrutilipi dataset}
We apply the above method to the AIR archive and extract 6,457 hours of data across 12 languages as detailed in Table~\ref{tab:aligned_data}, a yield of 67\% of all audio in the archive.
The data corresponds to 4.95M utterances with an average length of 8.9 words per sentence.
In Figure \ref{fig:existing-data}, we compare Shrutilipi to existing open-source public datasets as documented in the 2021 report \cite{giz} and Kathbath dataset. 
Shrutilipi increases the amount of labelled data by 2.3$\times$ on average across the 12 languages.  


\begin{table}[]
\centering
\small
\begin{tabular}{cccccc}
\toprule
\textbf{Lang.} & \textbf{\# S} & \textbf{\# ST} & \textbf{\# SN} & \textbf{\# A} \\
\midrule
bn & 4 & 12 & 360 & 2 \\
gu & 3 & 9 & 270 & 2 \\
kn & 3 & 9 & 270 & 2 \\
hi & 11 & 33 & 990 & 5 \\
ml & 3 & 9 & 270 & 2 \\
mr & 5 & 15 & 450 & 3 \\
or & 2 & 6 & 180 & 1 \\
pa & 1 & 3 & 90 & 1 \\
sa & 1 & 3 & 90 & 1 \\
ta & 4 & 12 & 360 & 2 \\
te & 4 & 12 & 360 & 2 \\
ur & 4 & 12 & 360 & 2 \\
\midrule
all & 45 & 135 & 4050 & 21 \\
\bottomrule
\end{tabular}
\caption{Statistics of human evaluation sampling (\# S = Stations; \# ST = Strata; \# SN = Sentences; \# A = Annotators)}
\label{tab:human_eval_stats}
\end{table}

\section{Evaluation of Shrutilipi}

In this section, we evaluate Shrutilipi along three axes:
(i) is the data of good quality?
(ii) is the data diverse? and
(iii) is it effective on downstream ASR?

\subsection{Is the data of good quality?}

We perform a human evaluation of Shrutilipi with data sampled across languages, regions, and alignment quality.


\paragraph{Annotation setup} 
The task is to check the quality of a mined audio and text pair with two Yes/No questions.
First, evaluators were shown the text and were asked if there were any mistakes in the text. 
This is to capture potential errors from the text processing pipeline. 
Then, evaluators listened to the audio and were asked if the audio aligns with the text. 
If they answered `No', they were asked to localize the error to (a) Start, (b) In Between, and (c) End of sentence. 
We built a custom web-interface for this task using LabelStudio \cite{labelstudio}.

Given the large variation we observe in the datasets across stations (speakers, content, and transcript formats), we sample data uniformly across the 45 stations across all languages.
We also sample data for different alignment scores ($\Delta$). 
Given that Shrutilipi is created with the threshold $\tau = 0.8$, we consider three intervals of the alignment score $\{[0.8-0.9), [0.9 - 0.95), [0.95, 1]\}$.
For each combination of the 45 stations and 3 score intervals, we uniformly sample 30 audio-text pairs, creating an annotation dataset with 4,050 items.

We recruited 21 human evaluators across the 12 languages.
The evaluators were native speakers in the respective language, and worked as full-time professional translators at a university. 
The evaluators were introduced to the task along with examples of potential errors they may expect in the data.






\begin{figure}[t]
    \centering
    \includegraphics[width=\columnwidth]{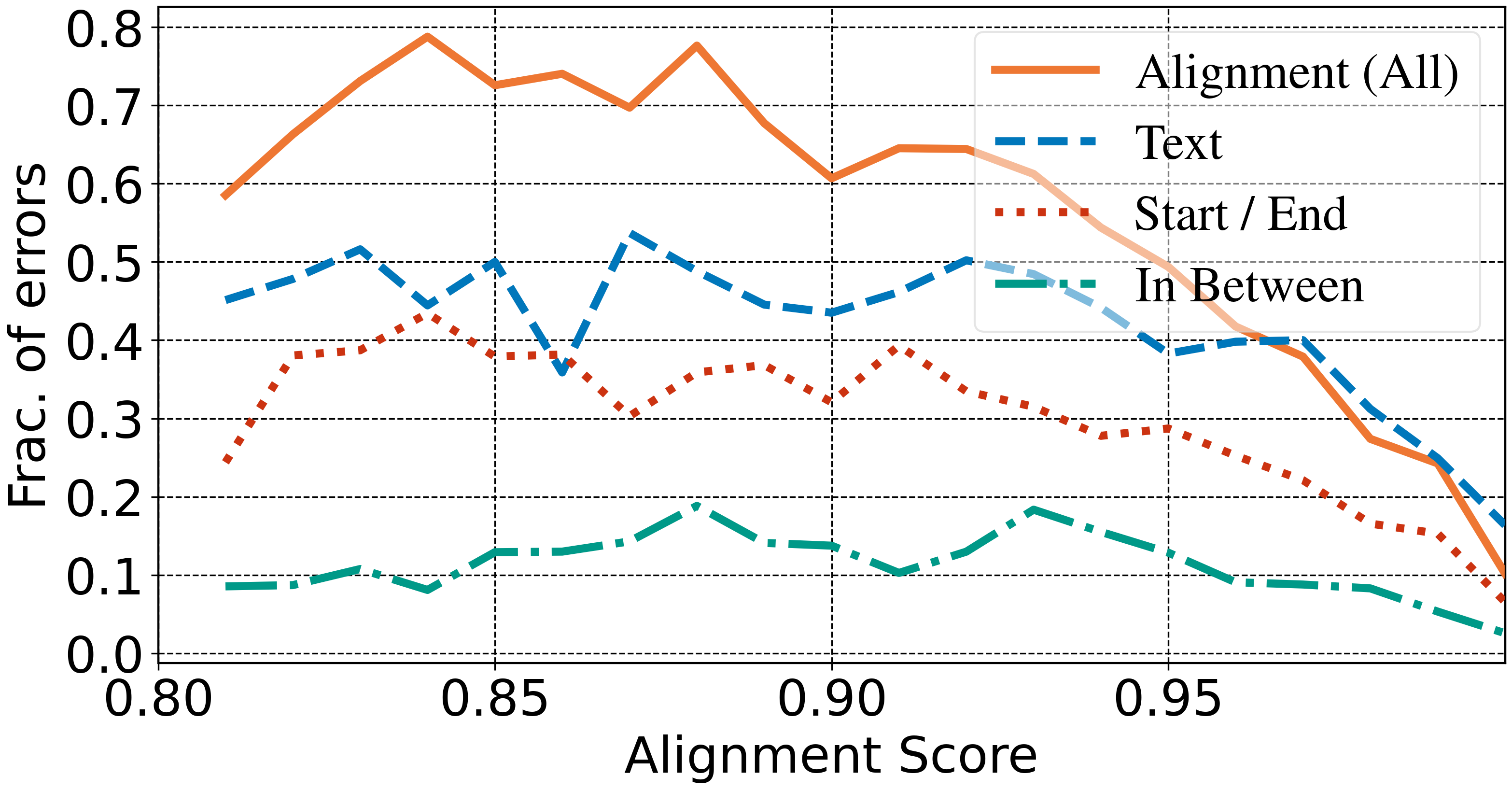}
    \caption{Fractions of annotation errors across languages for bucketed for different Alignment scores $\Delta$. }
    \label{fig:alignment-errors}
\end{figure}

\paragraph{Observations}
We summarize the results on the effect of variation of alignment score.
In Figure~\ref{fig:alignment-errors}, we plot the fractions of responses to different questions against different values of alignment score. 
We make three observations.
First, as expected, the fraction of errors reduces as alignment scores increase, with a marked reduction around the value of $0.95$.
Second, a large fraction of the errors (in the range $[0.8, 0.9]$) are due to errors in the original text, indicating the need for more accurate OCR and document understanding for Indian languages. 
Third, when localizing the error in alignment, a majority of the errors seem to be at the start or end of the audio segments.
Only a smaller fraction of errors are due to alignment issues within the audio segment, which incidentally do not show a strong dependence on the similarity score. 
We hypothesize training methodologies for E2E ASR systems would be forgiving of such errors.
In summary, the subset of Shrutilipi with 3,239 hours for similarity score $>$ 0.95, is rated to be of high accuracy, and for similarity score in $[0.8, 0.95]$ errors are primarily in the start or end of audio segments and often due to challenges in text extraction.

\subsection{Is the data diverse?}

     
     

\begin{figure}[t]
     \centering
     \includegraphics[width=\columnwidth]{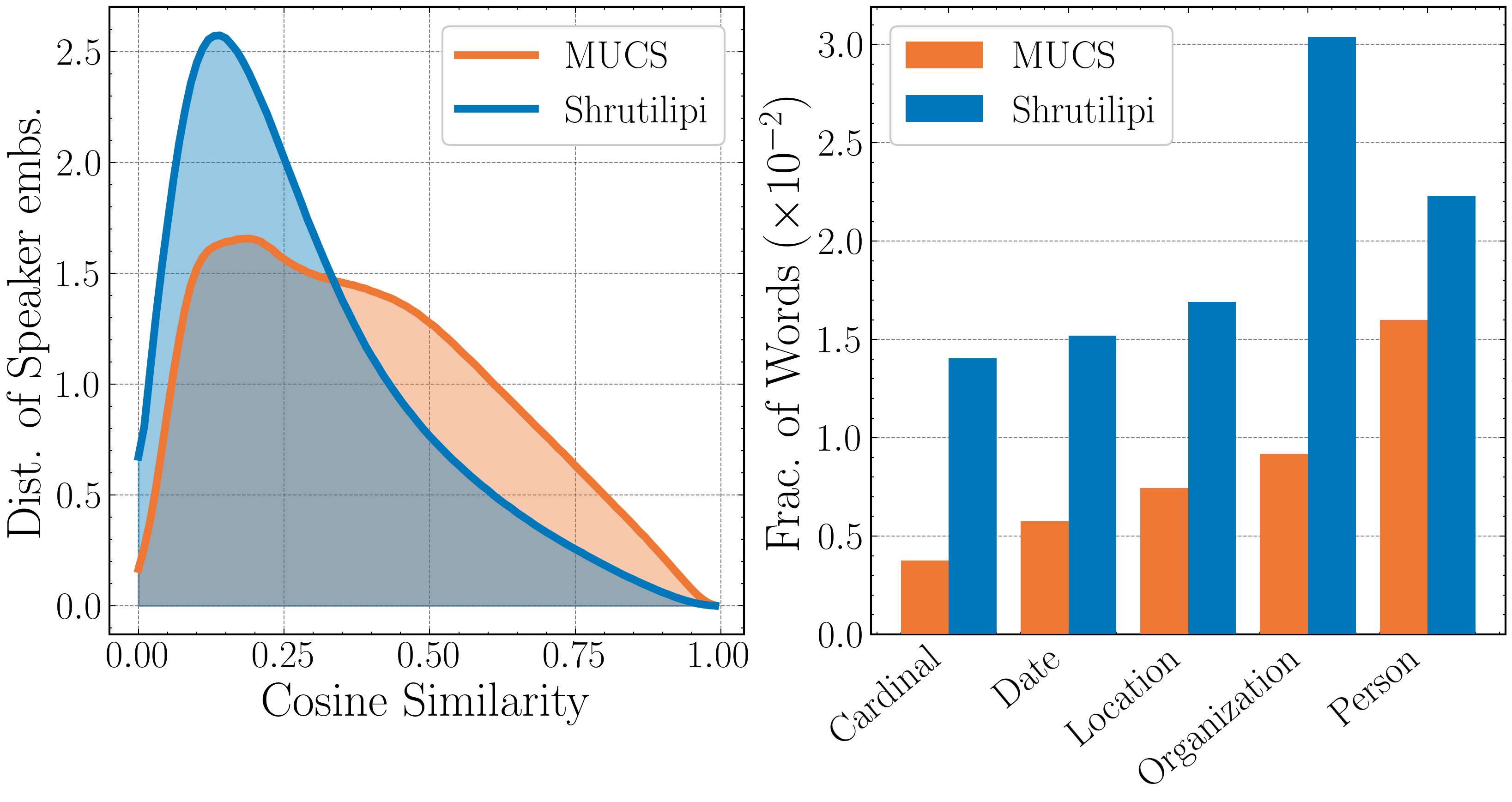}
          \caption{Diversity in Shrutilipi compared to MUCS}
        \label{fig:diversity}
     
\end{figure}



A key metric for labelled audio datasets is the diversity of speaker and content representation \cite{commonvoice}.
We compute metrics of diversity and compare against another publicly available dataset - MUCS \cite{mucs}.


\subsubsection{Diversity in Speakers} 
Increasing speaker diversity remains a key, and expensive, problem to solve.
The AIR dataset lends an opportunity to inexpensively mine diverse data from diverse regions.
To quantify speaker diversity, we build an Automatic Speaker Verification model to obtain speaker-specific embeddings. 
Specifically, we use the X-Vector model \cite{snyder2018x}, trained on Kathbath \cite{javed2022indicsuperb}.
We randomly sample 10K pairs of audio segments from the Hindi train sets of Shrutilipi and MUCS, and compute the cosine similarity of these pairs. 
We plot in Figure \ref{fig:diversity}, the distribution of the cosine similarity scores. 
The distribution for Shrutilipi denotes a much larger fraction of smaller similarity scores, indicating larger diversity. 


\subsubsection{Diversity in Named Entities}
An important metric for ASR systems is the performance on source-native named entities.
Current approaches of improving accuracy for named entities include integrating domain-specific external language models \cite{kannan2018analysis}, using contextual biasing \cite{pundak2018deep}, or boosting hotwords\footnote{https://github.com/kensho-technologies/pyctcdecode}. 
A more robust approach would be to collect datasets that represent diverse named entities.
To quantify this diversity, we count the number of occurrences of named entities in the Hindi datasets of Shrutilipi and MUCS. 
We translate the sentences to English using IndicTrans \cite{ramesh2022samanantar}, and then use Spacy's Entity Recognizer \cite{Honnibal_spaCy_Industrial-strength_Natural_2020} to obtain named entities.
We observe that MUCS and Shrutilipi datasets contain 766 and 222K unique named entities respectively, i.e., Shrutilipi provides a 290$\times$ increase. 
Further, the fraction of words that are named entities is also much larger in Shrutilipi across entity types (Figure~\ref{fig:diversity}). 


\subsection{Is it effective on downstream ASR?}

We evaluate the effectiveness of Shrutilipi as a training dataset for ASR systems. 
We consider both a large model - Wav2Vec \cite{baevski2020wav2vec}, and an efficient model - Conformer \cite{gulati2020conformer}.
We also create and test performance on a harder noisy benchmark.

\subsubsection{Models and Training} 

In this section we discuss the details of the two ASR Models - (i) Wav2Vec and (ii) Conformer.

\paragraph{Wav2Vec Model}

\subparagraph{Model Details} For all our experiments, we use the Wav2Vec \cite{baevski2020wav2vec} LARGE model consisting of 317M parameters. The model consists of 3 components - (i) a convolutional encoder, (ii) transformer blocks, (iii) and a linear projection head.  The convolutional encoder contains 7 convolutional layers each with 512 channels, strides of (5,2,2,2,2,2,2) and kernel widths of (10,3,3,3,3,2,2). The model has 24 transformer blocks with model dimension 1024 and FFN dimension 4096 with 16 attention heads. The linear projection head maps the output from the transformer block to the label set $L' = L \cup \{blank\}$, where $L$ is the set of all characters in the langauge. We initialize the model using the pretrained checkpoint from  \citet{javed2022towards}, which is pretrained on 17,000 hours of raw audio data across 40 Indian languages.

\subparagraph{Details of Finetuning} During finetuning, we use the Adam optimizer with a learning rate of $10^{-4}$ and a tri-stage learning schedule; linear warm-up for first 10\% of the steps, then held constant
for the next 40\% steps, and exponentially decayed for the remaining steps. We freeze the parameters of the convolutional encoder during fine-tuning. Additionally, we only update the parameters of the linear projection head for the first 200 steps. We train for 120K steps. We use the code from IndicWav2Vec\footnote{https://github.com/AI4Bharat/IndicWav2Vec} for finetuning.

\subparagraph{Details of Language Model}
We train 6-gram statistical language models for all 12 languages using KenLM library \cite{heafield-2011-kenlm} on IndicCorp dataset \cite{kakwani2020indicnlpsuite}. Before training, we clean the corpus by removing all those sentences which contain one or more characters that do not belong to the language, ensuring both the acoustic and language model has exactly same set of characters. We then augment it with training transcripts of respective ASR datasets leaving us with a total of 44M, 51M, 16M, 68M, 70M, 53M, 7M, 29M, 15M, 26M, 52M and 2M sentences for Bengali, Gujarati, Hindi, Kannada, Malayalam, Marathi, Odia, Punjabi, Sanskrit, Tamil, Telugu and Urdu respectively. Next, we train language models of order 6 and filter it using a custom lexicon created by choosing top 500K most frequent words in the training data. We also quantize all the n-gram probabilities (except unigrams) to 8 bits for faster inference.

During evaluation, we use a beam-search decoder along with the trained language model to decode the emissions from the softmax layer of acoustic model, using Flashlight\footnote{https://github.com/flashlight/flashlight} library, according to Equation \ref{eq:decoding}. 
\begin{equation}
\label{eq:decoding}
    \mathbf{y^*} = argmax_{\mathbf{y}} \log p_{AM}(\mathbf{y}) + \alpha \log p_{LM}(\mathbf{y}) + \beta |\mathbf{y}|
\end{equation}
where $|\mathbf{y}|$ is the length of the sequence and $\alpha$ and $\beta$ are hyperparameters. We set $\alpha$ and $\beta$ to 2 and -1 respectively and use a beam size of 128.   

\begin{table}[t]
\centering
\small
\begingroup
\setlength{\tabcolsep}{4.5pt} 
\renewcommand{\arraystretch}{0.7} 
\begin{tabular}{lcccccccc}
\toprule
 & bn & gu & hi & mr & or & ta & te & Avg. \\
\midrule
\multicolumn{9}{@{}l}{\textbf{MUCS Blind Set}} \\
\midrule
E & - & 17.9 & 12.0 & 13.6 & 23.3 & 20.5 & 16.4 & 17.3 \\
E+S & - & 12.8 & 11.1 & 11.4 & 23.0 & 20.7 & 13.8 & 15.5 \\
\midrule
\multicolumn{9}{@{}l}{\textbf{Kathbath Test Unknown}} \\
\midrule
E & 14.4 & 15.0 & 14.7 & 25.6 & 31.5 & 24.1 & 22.3 & 21.1 \\
E+S & 13.4 & 9.5 & 9.6 & 15.7 & 21.5 & 19.7 & 17.7 & 15.3 \\
\bottomrule
\end{tabular}
\endgroup
\caption{Results for Wav2Vec models on the Test Unknown of Kathbath trained on Existing and Shrutilipi datasets (E = Existing; S = Shrutilipi) }
\label{tab:indicsuperb}
\end{table}

\begin{table*}[t]
\centering
\small
\begingroup
\begin{tabular}{lcccccccc}
\toprule
Benchmarks & KB-K & KB-U & T & CV6 & CV7 & CV8 & CV9 & Avg. \\
\midrule
$M_{W2V}$ & 14.1 & 14.7 & 22.7 & 19.4 & 19.5 & 20.7 & 20.5 & 18.8 \\
$M+S_{W2V}$ & 9.4 & 9.6 & 19.7 & 15.0 & 13.4 & 13.9 & 13.7 & 13.5 \\
\midrule
$M +S_{\tau=0.95}$             & 9.8     & 10.2      & 20.4  & 16.4 & 14.2 & 14.9 & 14.7 & 14.4 \\
\midrule
$M_{Hard}$ & 19.2 & 17.9 & 26.5 & 22.6 & 24.4 & 25.8 & 25.8 & 23.2 \\
$M+S_{Hard}$ & 12.1 & 12.6 & 23.4 & 18.7 & 17.4 & 18.6 & 18.4 & 17.3 \\
\midrule
$M_{Conf.}$ & 17.2 & 17.7 & 25.4 & 20.9 & 21.4 & 22.9 & 22.8 & 21.2 \\
$M+S_{Conf.}$ & 15.2 & 14.9 & 23.9 & 19.3 & 19.1 & 20.0 & 19.9 & 18.9 \\
\bottomrule
\end{tabular}
\endgroup
\caption{Results on Hindi Benchmarks for Wav2Vec and Conformer models trained on MUCS and Shrutilipi datasets (M = MUCS; S = Shrutilipi; W2V = Wav2Vec; Hard = hard benchmark; Conf = Conformer; KB = Kathbath; K = Known; U = Unknown; T = Tarini; CV = CommonVoice)}
\label{tab:hindi_benchmarks}
\end{table*}



\paragraph{Conformer Model}

\subparagraph{Model details} For all our experiment, we use the Conformer \cite{gulati2020conformer} medium model consisting of 30.5M parameters. The model consists of 3 components - (i) a convolutional subsampling layer, (ii) conformer blocks, (iii) and a linear projection head. We extract 80-channel filterbanks features computed from a 25ms Hann window with a stride of 10ms from the raw audio. The convolutional subsampling layer has a stride of 4, transforming the 10ms frame rate to 40ms framerate. The model has 18 conformer blocks with model dimension 256 and FFN dimension 1024 with 4 attention heads. The linear projection head is similar to that of the Wav2Vec model, where the label set $L$ is created by tokenizing the data using Byte-Pair Encoding (BPE) \cite{sennrich-etal-2016-neural} with a vocab size of 128. We initialize our models from the pretained checkpoint from NGC\footnote{https://catalog.ngc.nvidia.com/orgs/nvidia/teams/nemo/models/\\ stt\_hi\_conformer\_ctc\_medium}, which is pre-trained on the ULCA\footnote{https://github.com/Open-Speech-EkStep/ULCA-asr-dataset-corpus} Hindi labelled dataset.

\subparagraph{Details of Finetuning} During finetuning, we use the Adam optimizer and the Noam Annealing schedule \cite{vaswani2017attention} with 1000 warm-up steps and peak learning rate of $2/\sqrt{d}$ where $d$ is the model dimension of the conformer block. We apply dropout \cite{srivastava2014dropout} in each residual unit of the conformer, with a rate of $P_{drop}$ = $10^{-3}$. We train for 50 epochs. We use the NeMo\footnote{https://github.com/NVIDIA/NeMo} libarary for finetuning.
We use 8 A100 GPUs for training all ASR models.

\subsubsection{Evaluation on multilingual benchmarks} 
We evaluate performance of Wav2Vec models on the blind set of MUCS \cite{mucs} and Test Unknown set of Kathbath \cite{javed2022indicsuperb}, as shown in Table \ref{tab:indicsuperb}. 
The model for Bengali was trained on the OpenSLR \cite{shetty2021exploring} train set, while we use the MUCS train set for other languages, denoted by E (Existing).
For the MUCS blind set, the average WER drops from 17.3\% to 15.5\%.
For Kathbath too, we see a large improvement of 5.8\% WER on average.



\subsubsection{Evaluation on Hindi benchmarks} 
Hindi has 7 benchmarks: Test Unknown and Test Known sets of Kathbath \cite{javed2022indicsuperb}, Tarini (not publicly available but shared privately upon request), CommonVoice \cite{commonvoice} versions 6, 7, 8 and 9. 
We evaluate Wav2Vec models trained on the MUCS \cite{mucs} train set and MUCS+Shrutilipi for Hindi, as shown in Table \ref{tab:hindi_benchmarks}. 
We see a consistent improvement in WER across all the 7 benchmarks, with an average improvement of 5.3\%. 
We also see that the model trained on Shrutilipi performs better than Shrutilipi with $\tau=0.95$, as seen in rows 2 and 3 of Table \ref{tab:hindi_benchmarks}. 

\subsubsection{Evaluation on efficient models} 
We train the Conformer model on MUCS train set and MUCS+Shrutilipi, and evaluate on the Hindi Benchmarks. 
Again, we see consistent improvement in WER across all benchmarks, wherein the Average WER improves from 21.2\% to 18.9\%, as seen in rows 6 and 7 of Table \ref{tab:hindi_benchmarks}.

\subsubsection{Evaluation on a hard benchmark} 
To evaluate if addition of Shrutilipi to the training set makes the models more robust to noise, we create a hard ASR benchmark for Hindi by adding background noise of various types to the audio files of the Hindi Benchmarks. 
Specifically, we use ESC dataset \cite{esc}, which consists of 2,000 short clips of background noise from 5 different categories. 
For each audio, we randomly pick a background clip and add it to the audio signal with a random Signal-to-Noise Ratio (SNR) value between 3 dB and 30 dB to control the intensity of noise added. 
We evaluate the Wav2Vec models trained on MUCS and MUCS+Shrutilipi for Hindi on the hard benchmark, as shown in rows 4 and 5 in Table \ref{tab:hindi_benchmarks}. 
There is an increase in WER values for all datasets and both models on the hard benchmark compared to the Hindi benchmark. 
On average, addition of Shrutilipi reduces WER by 5.9\%, a higher difference than with Hindi benchmark.

\section{Conclusion}

We consider creation of speech datasets from diverse publicly available datasets from All India Radio (AIR).
Given irregularities in data, we present a technique to mine audio and text-pairs at document scale by using CTC-based ASR models and the Needleman-Wunsch algorithm. 
By applying this technique on the AIR archives, we create the Shrutilipi dataset, which consists of 6,457 hours of labelled audio for 12 Indian languages.
We show that Shrutilipi is of good quality and has significantly higher diversity in speakers and content in comparison to other public datasets. 
We evaluate its effectiveness on downstream ASR by evaluating on multiple benchmarks, training on efficient models, and showing robustness to noise. 
We hope that this methodology is applicable to other public datasets and other languages as well to advance speech technology for low-resource languages.

\section*{Acknowledgements}

We would like to thank the Ministry of Electronics and Information Technology (MeitY\footnote{https://www.meity.gov.in/}) of the Government of India and the Centre for Development of Advanced Computing (C-DAC\footnote{https://www.cdac.in/index.aspx?id=pune}), Pune for generously supporting this work and providing us access to multiple GPU nodes on the Param Siddhi Supercomputer. We would like to thank the EkStep Foundation and Nilekani Philanthropies for their generous grant which went into hiring human resources as well as cloud resources needed for this work. We would like to thank Megh Makhwana from Nvidia for helping in training Conformer-based ASR models. We would like to thank the EkStep Foundation for providing the Tarini dataset. We would like to thank Janki Nawale and Anupama Sujatha from AI4Bharat for helping in coordinating the annotation task, and extend thanks to all the annotators of AI4Bharat team. 
\bibliography{anthology,custom}
\bibliographystyle{acl_natbib}




\end{document}